\documentclass[10pt,twocolumn,letterpaper]{article}

\usepackage{wacv}
\usepackage{times}
\usepackage{epsfig}
\usepackage{graphicx}
\usepackage{amsmath}
\usepackage{amssymb}
\usepackage{multirow}
\usepackage{subfigure}
\usepackage{color}
\usepackage{enumitem}

\wacvfinalcopy 

\usepackage[breaklinks=true,bookmarks=false]{hyperref}

\begin{document}

\newcommand{\ba}{\mbox{\boldmath $a$}}
\newcommand{\bb}{\mbox{\boldmath $b$}}
\newcommand{\bc}{\mbox{\boldmath $c$}}
\newcommand{\bd}{\mbox{\boldmath $d$}}
\newcommand{\bdh}{\mbox{\boldmath $\hat{d}$}}
\newcommand{\bt}{\mbox{\boldmath $t$}}
\newcommand{\bw}{\mbox{\boldmath $w$}}
\newcommand{\bx}{\mbox{\boldmath $x$}}
\newcommand{\R}{\mathbb R}
\newcommand{\bphi}{\mbox{\boldmath $\phi$}}
\newcommand{\ua}{\uparrow}

\title{Coupled Depth Learning}

\author{Mohammad Haris Baig\\
Dartmouth College\\
Hanover, New Hampshire.USA\\
{\tt\small haris@cs.dartmouth.edu}
\and
Lorenzo Torresani\\
Dartmouth College\\
Hanover, New Hampshire.USA\\
{\tt\small lt@dartmouth.edu}
}
\maketitle

\begin{abstract}
In this paper we propose a method for estimating depth from a single image using a coarse to fine approach. We argue that modeling the fine depth details is easier after a coarse depth map has been computed. We express a global (coarse) depth map of an image as a linear combination of a depth basis learned from training examples. The depth basis captures spatial and statistical regularities and reduces the problem of global depth estimation to the task of predicting the input-specific coefficients in the linear combination. This is formulated as a regression problem from a holistic representation of the image.  Crucially, the depth basis and the regression function are {\bf coupled} and jointly optimized by our learning scheme. We demonstrate that this results in a significant improvement in accuracy compared to direct regression of depth pixel values or approaches learning the depth basis disjointly from the regression function. The global depth estimate is then used as a guidance by a local refinement method that introduces depth details that were not captured at the global level. Experiments on the NYUv2 and KITTI datasets show that our method outperforms the existing state-of-the-art at a considerably lower computational cost for both training and testing.
\end{abstract}

\section{Introduction}
Over the last few years depth estimation has been the subject of active research by the machine learning and computer vision community~\cite{BaigJPBDS14, Karsch:TPAMI:14,konrad2013learning,Liu_2014_CVPR,Ladicky_2014_CVPR}. This can partly be attributed to the fact that algorithms using the depth channel as an additional cue have shown dramatic improvements over their RGB counterparts on a number of challenging vision problems~\cite{ZhangICCV2013,guptaECCV14,Hen12RGB,Hermans14ICRA}. Most of these improvements have been demonstrated using depth measured by hardware sensors. However, most of the pictures available today are still traditional RGB (rather than RGBD) photos. Thus, there is a need to have robust algorithms for estimating depth from single RGB images.

While inferring depth from a single view is ill-posed in general (an infinite number of 3D geometric interpretations can fit perfectly well any given photo), physical constraints and statistical regularities can be exploited to learn to predict depth from an input photo with good overall accuracy.
In this work we propose to learn these spatial and statistical regularities from a RGBD training set in the form a global depth basis. We hypothesize that the depth map of any image can be well approximated by a linear combination of this global depth basis. Following this reasoning we formulate coarse depth estimation as the problem of  predicting the coefficients of the linear combination from the input image. Our design choice makes this regression problem easier as the target dimensionality is much lower than the number of pixels and the output space is more structured. Crucially, we learn the depth basis and the regression model {\em jointly} by optimizing a single learning objective. We denote our global estimation method as {\bf GCL} (Global Coupled Learning).

As input for our regression problem we use a holistic image representation capturing the coarse spatial layout of the scene. While in principle we could attempt to learn this holistic feature descriptor too, we argue that existing RGBD repositories are too limited in scope and size to be able to learn features that would generalize well to different datasets. Instead, we propose to leverage a pretrained global feature representation that has been optimized for scene classification~\cite{zhou2014places}. The intuition is that since these features have been tuned to capture spatial and appearance details that are useful to discriminate among a large number of scene categories, we expect them to be also effective generic features for the task of depth prediction. Our experiments on two distinct benchmarks validate this hypothesis, showing that our models trained on these scene features yield state-of-the-art results (without any fine-tuning). 

%

\newcommand{\mywidth}{0.20}
\begin{table*}[!htbp]
{\footnotesize
    \begin{center}
    \begin{tabular}{|c|c|c|c|c|}
	\hline
    	Input Image & Ground Truth & GCL (global)  & RCL (refinement) \\
	\hline
	 & & & \\
      	\includegraphics[width=\mywidth\linewidth]{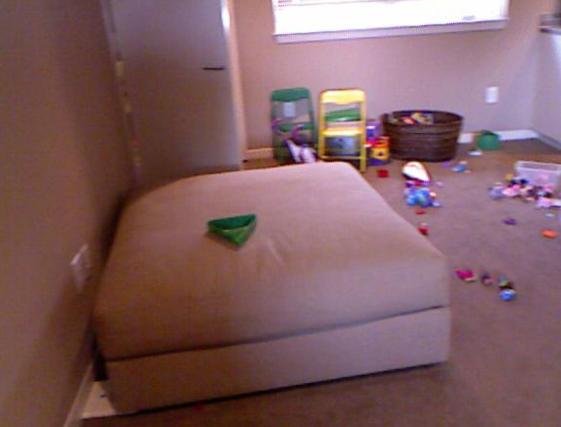} & \includegraphics[width=\mywidth\linewidth]{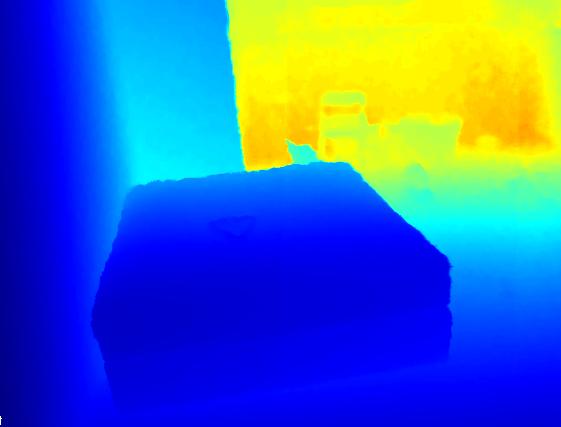}
      	& \includegraphics[width=\mywidth\linewidth]{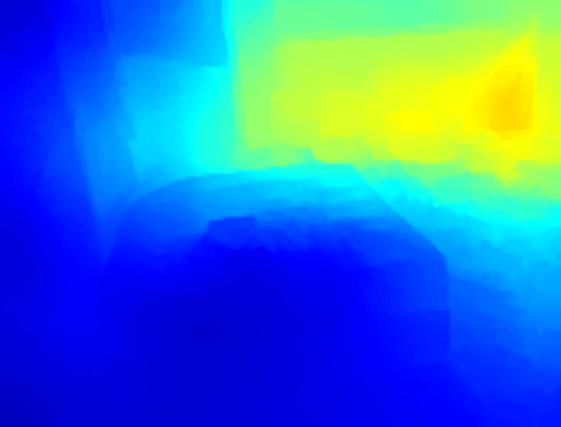} & \includegraphics[width=\mywidth\linewidth]{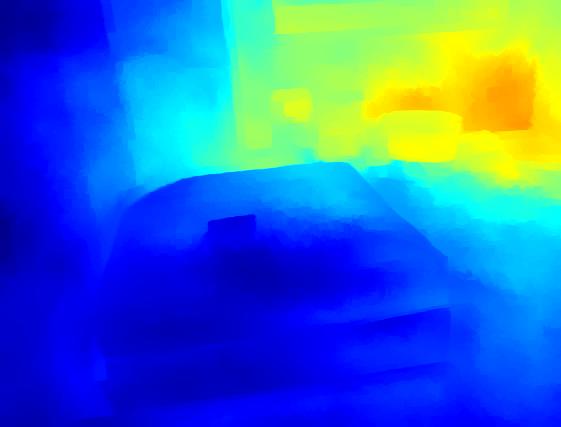}   \\
	      	\includegraphics[width=\mywidth\linewidth]{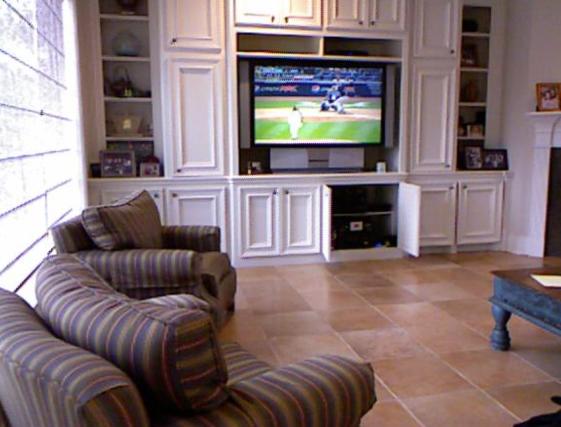} & \includegraphics[width=\mywidth\linewidth]{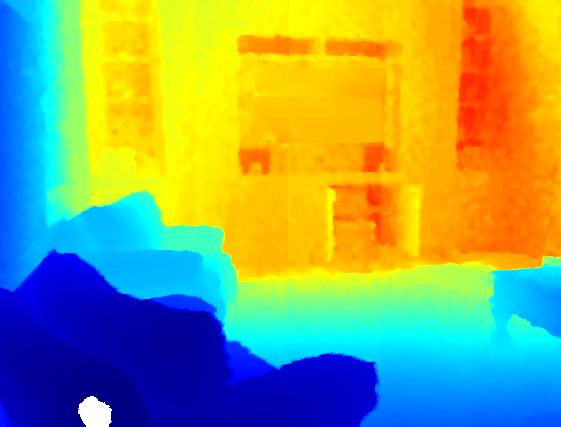}
      	& \includegraphics[width=\mywidth\linewidth]{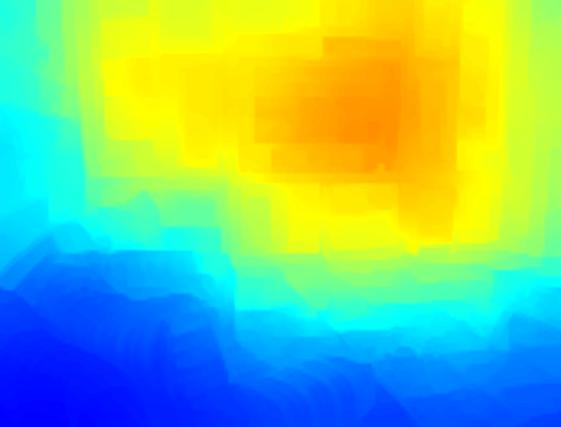} & \includegraphics[width=\mywidth\linewidth]{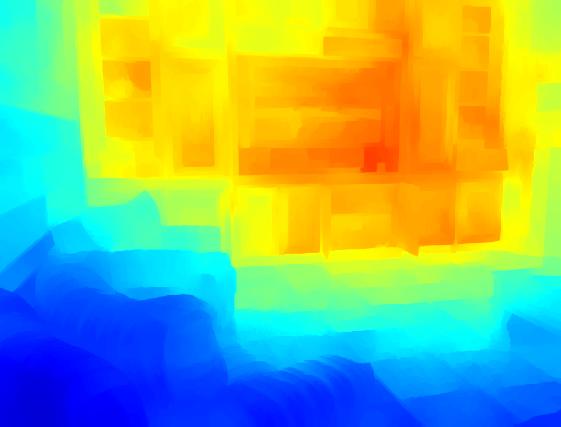}  \\
      	\includegraphics[width=\mywidth\linewidth]{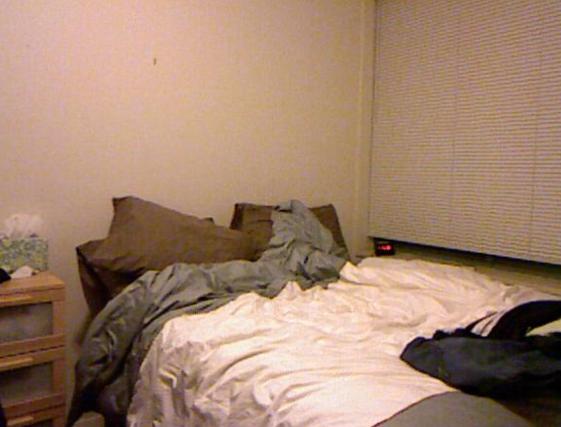} & \includegraphics[width=\mywidth\linewidth]{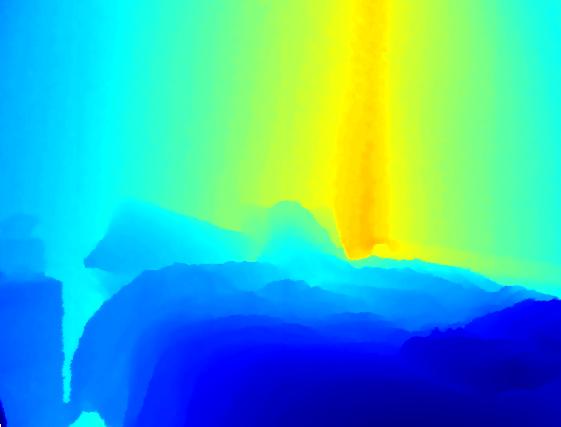}
      	& \includegraphics[width=\mywidth\linewidth]{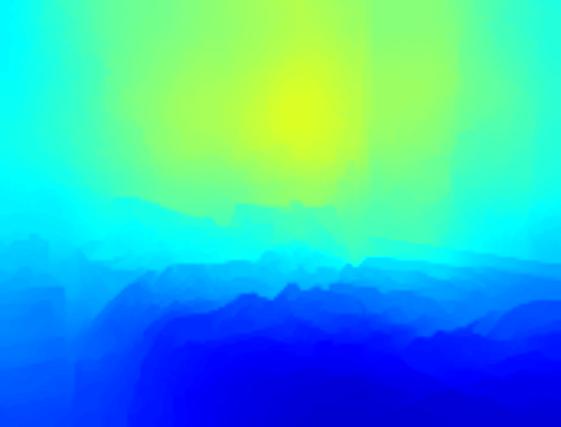} & \includegraphics[width=\mywidth\linewidth]{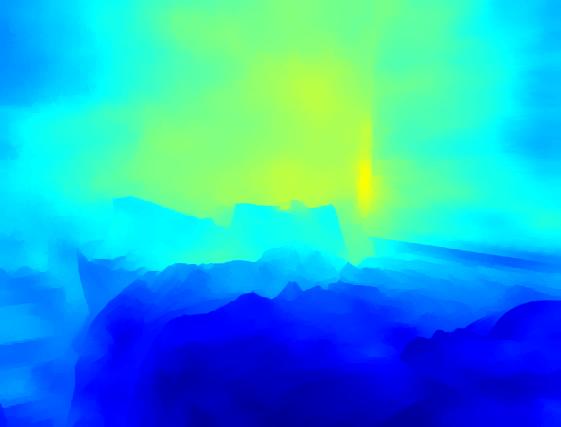}  \\
%
      	\hline
    \end{tabular}
    \end{center}
    }
    \caption{Visualization of depth estimates from NYUv2. Notice how the local refinement (RCL) captures finer depth details compared to GCL, such as the corner of the bed and the objects in the background of the first photo example, or the bookcase in the second picture or the object on the bed and the corner of the room in the third picture.}
\label{tb:visResultsNYUV2}
\end{table*}

Since our model is trained on a holistic description of the image, it can be argued that it is implicitly optimized to predict the main global 3D structure in the scene, possibly at the expense of fine depth details. To address this potential shortcoming we propose a local refinement step, which uses the global estimate to improve the depth prediction at individual pixels. We refer to this refinement procedure as {\bf RCL} (Refined Coupled Learning). This is achieved by training a depth refinement function on hypercolumn features~\cite{BharathCVPR2015} of individual pixels, which describe the local appearance and context in the neighborhood of the pixel. Our experiments indicate that the local refinement quantitatively improves the global estimate and produces finer qualitative details. In Table~\ref{tb:visResultsNYUV2} we show the global (GCL) and locally-refined (RCL) depth outputs produced by our system for a few example  images.

\section{Related Work}

While initial approaches to depth estimation exploited specific cues like shading~\cite{Zhang99shapefrom} and geometry~\cite{Hedau_2009_SpatialLayout},
more recently the focus has shifted toward employing pure machine learning methods due to the heavily restrictive assumptions of these earlier methods. Most of the earlier machine learning based approaches~\cite{saxena2005learningdepth, saxena2009make3d, LiuCVPR2010semanticdepth} operate in a bottom-up fashion by performing local prediction (e.g., estimating depth for individual patches or superpixels) and by spatially smoothing these estimates with a CRF or a MRF model.
The advantage of local prediction models is that they can be trained well even with limited RGBD data since they treat each patch or pixel in the collection as a separate example. However, small regions do not capture enough context for robust depth estimation. In contrast, we approach depth estimation first at a global level by considering the entire image at once. Then we regress depth at a per-pixel level using the global estimate as a prior.

With the advent of larger RGBD repositories~\cite{Silberman:ECCV12,SUN3D} there has been an increased interest in the use of nonparametric methods~\cite{Karsch:TPAMI:14, konrad2013learning} for depth estimation.
These approaches find nearest-neighbors of the query in the training set, and then fuse the depth maps of the retrieved neighbors to produce a depth estimate for the query. Such approaches do not generalize well unless the test set is collected in the same exact environment as the training set. This imposes large computational and memory constraints on their usage. 

Recently, there has been an increased interest in applying deep learning methods~\cite{EigenPF14,CVPR15bFayao,Wang_2015_CVPR} for estimating depth from a single image. Most of these systems~\cite{EigenPF14,Wang_2015_CVPR} attempt to regress depth directly from the image. This requires learning large models that can be trained effectively only with hundreds of thousands of examples. This renders these techniques inapplicable in areas where data is scarce. Liu et al.~\cite{CVPR15bFayao} proposed learning deep features for super-pixel depth prediction. A CRF is then applied to enforce global coherence over the entire depth map. While this approach does work for smaller datasets, it is restricted to perform coarse super-pixel predictions where each super-pixel is assumed to be facing the camera (has no depth gradient).

Our approach critically differs from prior work in two fundamental aspects. First, our approach predicts a small set of depth reconstruction weights rather than the full depth maps. Our design choice exploits statistical regularities in the problem and reduces the number of outputs to predict. We demonstrate that this allows our method to achieve a much lower RMSE error than methods predicting depth maps directly~\cite{EigenPF14,Wang_2015_CVPR,CVPR15bFayao}, even when using 150 times less training data (on NYUv2).  Second, our refinement model is trained to predict the depth at individual pixels using local pixel descriptors rather than super-pixels~\cite{CVPR15bFayao}. Furthermore, we also show how to leverage features from deep networks trained on related tasks to further improve performance.

\begin{figure*}[t]
\begin{center}
(a)\includegraphics[height=0.15\linewidth]{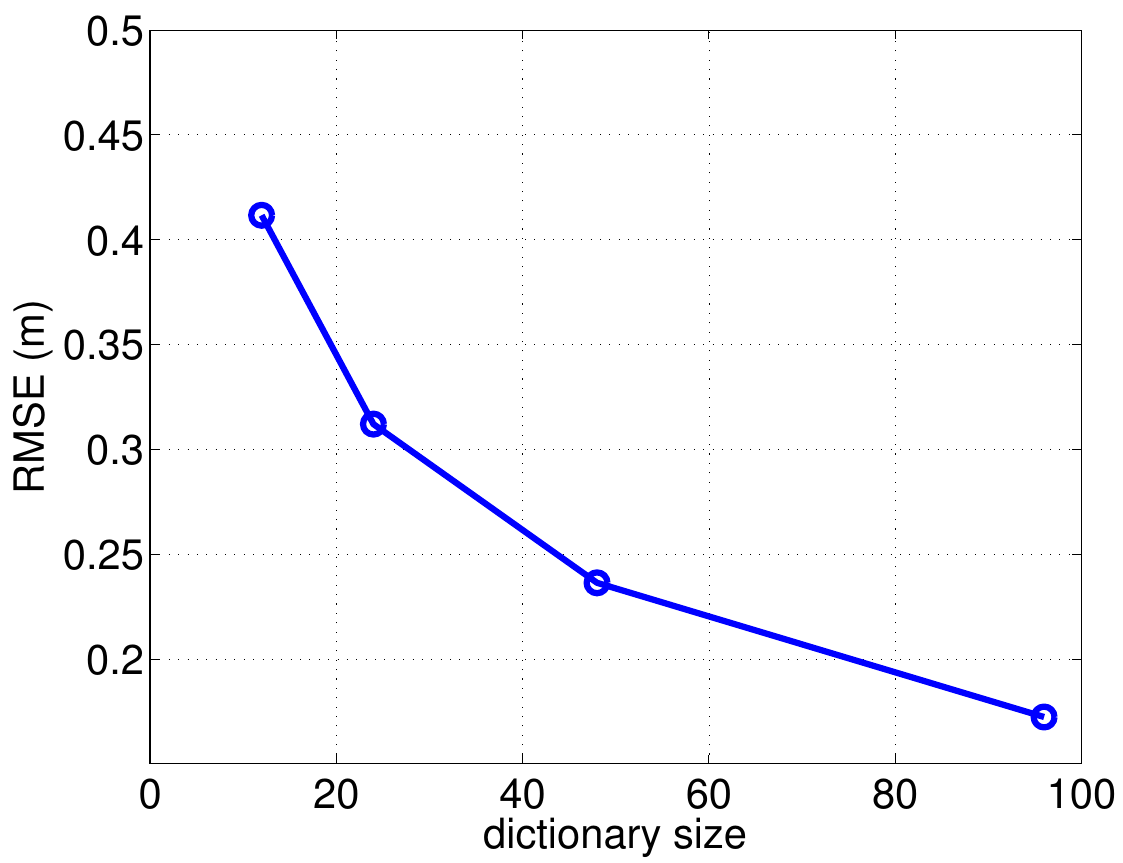}~~(b) \includegraphics[height=0.14\linewidth]{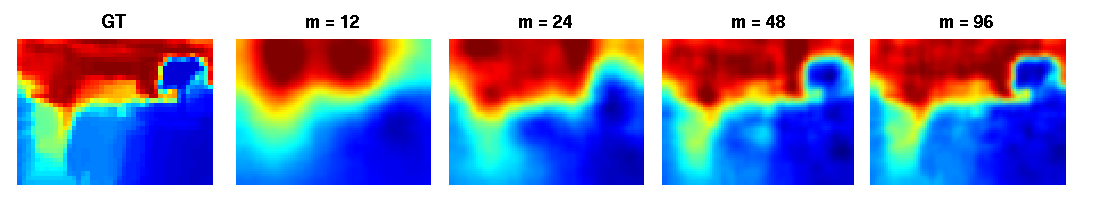}
\end{center}
   \caption{{(a) Reconstruction error (RMSE) of ground truth depth on NYUv2 for different dictionary sizes (note that this experiment did not involve depth prediction from images, just ground truth depth approximation). For NYUv2 we use a dictionary of size 48 as this provides a good compromise in terms of compactness and approximation quality. (b) Qualitative effects of different dictionary sizes on a sample depth map. From left to right: ground truth (GT) and least-square approximations using a dictionary with size $m=12, 24, 48, 96$, respectively.}}
\label{fig:varyingDictionary}
\end{figure*}

Our joint optimization of depth basis and regression is inspired by prior work in semi-coupled dictionary learning~\cite{wang2012semi}. Here we borrow this optimization scheme to perform joint learning of a depth dictionary and a regressor from the image space to the basis weights in order to reduce the number of outputs to predict. While we focus on the the problem of depth estimation from single view, we believe that our approach can be used effectively in other scenarios involving dense pixel-level predictions under limited availability of training data.

\section{Technical Approach}

In the following subsections we discuss how to \textbf{jointly} learn a global depth basis and a transformation from a given image space to the basis weights using training data. Then we discuss how to use the trained model to infer the coarse global depth of an image. Finally, we show how to further refine the coarse estimate with pixel-level predictions. 

Let $\mathcal{D} =\{(X_1, D_1), \hdots, (X_N, D_N)\}$ be the training set used to learn our model, where $X_i \in \R^{R \times C \times 3}$ represents the $i$-th image (consisting of $R$ rows, $C$ columns and 3 color channels) and $D_i \in \R^{R \times C}$ is its associated ground-truth depth map.



\subsection{Global Depth Estimation}

\subsubsection{Learning the Global Depth Model}
\label{coarsemodel}

To learn the global depth model, we start by downsampling the ground truth training depth maps. This has the effect of removing fine depth details (object boundaries, fine gradients denoting local shape, etc). We denote with $\bd_i \in \R^{P_L}$ the vector obtained by vectorizing the depth map $D_i$ after resizing to a lower resolution, where $P_L$ represents the dimensionality of the low resolution depth map. Similarly, we indicate with $\bx_i \in  \R^{R \cdot C \cdot 3}$ the vector obtained by stacking the pixel values of the image one on top of the other. Our objective is to train a model that, given an input image $\bx$ (at full resolution), predicts the global depth map $\bd$.

The first assumption we make is that the global depth map $\bd$ can be expressed as a linear combination of basis vectors from a depth basis $B = \left[\bb_1, \hdots, \bb_m\right]$:
\begin{equation}
\bd = B \bw
\label{eq:dict}
\end{equation}
where the $\bb_k \in \R^{P_L}$ are the basis atoms and $\bw = \left[w_1, \hdots, w_m\right]^\top$ is the vector containing the image-specific mixing coefficients (or weights). {Fig.~\ref{fig:varyingDictionary} shows both quantitatively as well as qualitatively the effect of varying the dictionary size on depth reconstruction.} We propose to learn a mapping $h: \R^{R \cdot C \cdot 3} \rightarrow \R^m$ that predicts the depth reconstructive weights $\bw$ from the input image $\bx$. Note that in our work $m << P_L$ (e.g., $m=48$ for NYUv2 and $m=96$ for KITTI) and thus the use of the depth basis greatly reduces the number of outputs that the regression model needs to predict. To regress on $\bw$ we choose a simple kernel-based regression model
\begin{equation}
h(\bx) = T \bphi(\bx) \approx \bw
\label{eq:regr}
\end{equation}
where $T \in \R^{m \times n}$ and $\bphi(\bx) = \left[\phi_1(\bx), \hdots, \phi_n(\bx)\right]^T$ is a vector containing  radial basis functions $\phi_j(\bx)$ computed with respect to centers $\bc_j$ for $j=1, \hdots, n$. The centers $\bc_1, \hdots \bc_n$ are example images (different from those included in the training set $\mathcal{D}$) and selected according to the details described in section \ref{implementationdetails}. Intuitively, they represent $n$ prototypical images that allow us to express $\bw$ as a linear combination of kernel distances from $\bx$. We compute the radial basis functions in terms of feature descriptors $f(\bx), f(\bc)$ extracted from the images $\bx, \bc$. We use as image representation $f(\bx)$ the features computed by layer ``$pool_5$'' of the deep network of the PLACES model~\cite{zhou2014places}. This is the max-pooled output of the fifth and final convolutional layer in the network. This feature map has dimensionality $6 \times 6 \times 256 = 9216$. While prior work~\cite{Girshick:CVPR14, Panda, Karayev} has shown that the subsequent (fully connected) layers of the Krizhevsky~\cite{Krizhevsky} network (same architecture, different dataset) produce higher level representations that yield improved recognition accuracy, $pool_5$ is the most appropriate feature map to use in our setting since it is the last layer preserving explicit location information before the ``spatial scrambling'' of the fully connected layers. Note that a spatially-{\em variant} representation is crucially necessary to predict the depth at each pixel. We validated experimentally this intuition and observed that using the feature maps from the fully-connected layers produced poorer depth prediction accuracy. Using this representation for feature vector $f(\bx)$, we then compute $\phi_j(\bx) = \exp(-||f(\bx) - f(\bc_j)||^2 / 2\sigma_{j}^2)$.

Given this model, a na\"{\i}ve approach to training our depth estimator is to learn disjointly the depth basis and the regression mapping. This would involve first learning the basis $B$ and the weights $\bw$ of Eq.~\ref{eq:dict} (e.g., by minimizing the reconstruction error on training depths $\bd_1, \hdots, \bd_N$) and then  regressing on these learned weights to estimate the transformation $T$ of Eq.~\ref{eq:regr}. While straightforward, in our experiments we demonstrate that this two-step process yields much inferior results compared to a {\em joint} optimization over $B$, $\bw$, and $T$ using a single learning objective that couples all of the parameters together. We refer to this learning objective as $J(B, \bw, T)$ and define it as follows:
\begin{eqnarray}
J(B, \bw, T) &=& \sum\limits_{i=1}^{N}|| \bd_i - B \bw_i ||_2 + \lambda_w \sum\limits_{i=1}^{N}||\bw_i||_1\\ \nonumber
  & & +~\lambda_{r} \sum\limits_{i=1}^{N} || \bw_i - T \bphi (\bx_i) ||_{2} + \lambda_{T}||T||_{F}~.
\label{eq:coarseobjective}
\end{eqnarray}
The first two terms of $J$ encourage reconstruction of the depth maps using {\em sparse} weights and are equivalent to the terms of the traditional sparse coding objective~\cite{Lee_etal07:sparseCoding}. The third term imposes the requirement that the depth weights be ``predictable'' under the regression model. The final term is a regularizer over the transformation $T$. Thus, joint optimization of $J$ over all parameters will yield a depth basis $B$, depth weights $\bw$, and transformation $T$  that simultaneously minimize 1) sparse reconstruction of depths maps and 2) regression error from the image domain to the depth space, subject to appropriate regularizations. In practice we minimize $J(B, \bw, T)$ with the added constraints $|| \bb_j ||_2 \leq 1$ for $j=1, \hdots, n$ in order to avoid scale degeneracies on  $B$. Furthermore, we enforce positivity constraints on the sparse weights $w_{ij}$ in order to define a purely additive depth model. We have found experimentally that this yields slightly better results than leaving the weights unconstrained. We also considered using an L2 sparsity over the weights $\bw_i$ but found that this produces consistently slightly worse results, as also reported in prior articles~\cite{ng2004feature}.

While our learning objective is not jointly convex over $\bw, B, T$, it is convex for each of these individual parameters when we keep the other two fixed. Based on this, we optimize our learning objective via block-coordinate descent by minimizing in turn with respect to 1) the basis, 2) the depth weights and 3) the transformation. These three alternating steps are discussed in detail below:
\begin{enumerate}[leftmargin=.75cm,nolistsep,align=left]
\item {\bf Estimate weights $\bw$ given parameters $B$, $T$.} It is easy to verify that minimizing $J$ with respect to $\bw$ while keeping $B$, and $T$ fixed (at the current estimate) reduces to a problem of the form:
\begin{equation}
\arg \min_{\bw} \sum\limits_{i=1}^{N}|| \ba_i -  C \bw_i ||_2 + \lambda_w \sum\limits_{i=1}^{N}||\bw_i||_1
\end{equation}
where $\ba_i, C$ are constants written in terms of $B$ and $T$. We solve this problem globally via least angle regression (LARS)~\cite{Efron04leastangle}.
\item {\bf Learning the depth basis $B$ given $\bw$, $T$}. This amounts to a L2-constrained least-squares problem, which we solve using the Lagrange dual, as in Lee et al.~\cite{Lee_etal07:sparseCoding}.
\item {\bf Learning the transformation $T$ given $\bw$, $B$}. This reduces to a L2-regularized least-squares problem, which can be solved in closed-form as shown in Wang et al.~\cite{wang2012semi}.\\
\end{enumerate}

\begin{figure}[t]
\begin{center}
\includegraphics[width=0.5\linewidth]{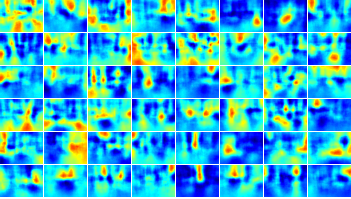}  ~~  \includegraphics[width=0.45\linewidth]{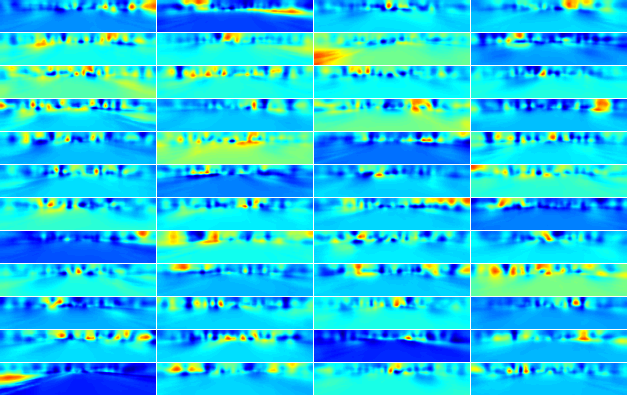}
\end{center}
   \caption{The depth basis learned by our global model GCL on the NYUv2 dataset (left) and on KITTI (right). The basis is used to model the structure of the output space (depth), thus reducing the complexity of the regression problem.}
\label{fig:coarsedict}
\end{figure}

We initialize this optimization by setting $B$ and $\bw$ to the solution computed via sparse coding~\cite{Lee_etal07:sparseCoding}, thus neglecting the terms in $J$ depending on transformation $T$. We then compute $T$ by solving step 3 above. Fig.~\ref{fig:coarsedict} shows the bases learned with this procedure on  NYUv2 and KITTI.

\subsubsection{Global Depth Map Inference}
\label{sec:globalinference}

At inference time, given a new input image $\bx$, we compute its global depth map $\bd^G$ by finding the {\em sparse} depth weights $\bw$ that best fit the image-based prediction, i.e., by solving the following optimization problem subject to positivity constraints on the weights:
\begin{equation}\label{eq:computeWeights}
    \arg\min_{\bw} || \bw - T \bphi(\bx) ||_{2} + \lambda_w ||\bw||_1~.
\end{equation}
The global depth map is then generated as $\bd^G = B \bw$.

Empirically, we have found beneficial to apply the colorization procedure described in Levin et al.~\cite{Dani04colorization} to the global estimate $\bd^G$ produced by our approach. This technique has been used in previous work~\cite{Silberman:ECCV12, EigenPF14} to fill-in missing values in data collected by depth sensors. Here instead we use it to make the depth map more spatially coherent, as the colorization procedure encourages pixels having similar color to be mapped to similar values of depth. To do this, we first resize the low-resolution depth map $\bd^G$ via bilinear interpolation to an intermediate size of $P_I$ pixels. 
Then we apply the colorization procedure using all pixels at this resolution as ``color'' propagation seeds with a low penalty value (the penalty value indicates how much the colorized depth values can deviate from the original input value). The details of this procedure are discussed in section \ref{implementationdetails}.

\subsection{Local Depth Refinement}

The local depth refinement uses the prediction $\bd^G$ from our global depth model (described in the previous section) and generates a higher resolution, locally-refined depth map $\bd^{R\ua}$ containing finer details. Let $\bd^{G\ua}$ be the global depth estimate resized to the intermediate resolution $P_I$ and post-processed via colorization as described in subsection~\ref{sec:globalinference}. Also, we denote with $\bd_i^\ua$ the ground truth depth map $D_i$ resized to size $P_I$ and vectorized. We propose to train a {\em local} refinement model that predicts the depth of pixel $j$ in example $i$ using a local descriptor $\bphi^\ua_j(\bx_i)$ computed at pixel $j$, i.e., $d_{ij}^\ua \approx \bt^{\ua} \cdot \bphi^\ua_j(\bx_i)$, where $\bt^\ua$ is a row vector encoding the model parameters. Note that this parameter vector is shared across pixels but, unlike our global depth estimator, the refinement predicts the depth at a pixel using as input a local descriptor computed at that pixel rather than the whole image. Specifically, we choose
\begin{equation}
\bphi^\ua_j(\bx_i) = \left[1~~d_{ij}^{G\ua}~~ {\phi}^\ua_{j1}(\bx_i), \hdots, {\phi}^\ua_{jn^\ua}(\bx_i)\right]^T
\label{eq:localfeature}
\end{equation}
where $d_{ij}^{G\ua}$ is the depth estimate for pixel $j$ in image $i$ from our global model, which is used as additional feature here in order to guide the local refinement. Thus, the global depth estimate acts in a sense as a prior for the local refinement, which lacks the context of the full-image. In our experiments we show that providing $d_{ij}^{G\ua}$ as feature is critically necessary to achieve good accuracy in the local refinement. The first feature entry is set constant to $1$ in order to implement the bias term. Finally, the features ${\phi}^\ua_{j1}(\bx_i), \hdots, {\phi}^\ua_{jn^\ua}(\bx_i)$ are radial basis functions computed with respect to $n^\ua$ centers. Note that while the radial basis functions for the global model were defined in terms of deep features $f(\bx)$ computed from the whole image, global features are clearly not appropriate for the local refinement. Instead, we propose to use the hypercolumn feature vector~\cite{BharathCVPR2015} at pixel $j$, i.e., the activation values at location $j$ in the convolutional feature maps of the deep PLACES network~\cite{zhou2014places} all stacked into a single vector. In practice, we use only layers {\em pool2, conv4 and conv5}, which give rise to a hypercolumn vector of dimensionality $896$ for each pixel. This representation has been shown to be able to simultaneously capture {\em localized} low-level visual information (from the early layers) as well as high-level semantics (from the deepest layers). Thus, it is very useful for localized, high-level visual analysis, such as our task of local depth refinement. More formally, we compute the radial basis features as ${\phi}^\ua_{jk}(\bx_i) = \exp(-||\alpha_j^\ua(\bx_i) - \bc^\ua_k||^2 / 2\nu^2)$ where $\alpha^\ua_j()$ denotes the function that extracts the hypercolumn representation at pixel $j$ and $\bc^{\ua}_k$ is the $k$-th center, itself a hypercolumn feature vector. As discussed in further detail in section~\ref{implementationdetails}, the centers $\bc^{\ua}_k$ are the centroids computed by k-means over a training set of hypercolumn feature vectors.

The parameter vector $\bt^\ua$ is learned via simple regularized least-squares estimation on the training data:
\begin{equation}\label{eq:computeLocalWeights}
    \arg\min_{\bt^\ua} \sum_{i=1}^N \sum_{j=1}^{P_I} \left(d_{ij}^\ua - \bt^{\ua} \cdot \bphi_j^\ua(\bx)\right)^2 + \lambda_t \sum_{k=3}^{n^\ua} (t^\ua_k)^2
\end{equation}
where the first two entries of $\bt^\ua$ (corresponding to the bias and the global depth prediction) are left unregularized.

At test time, given the input image $\bx$ {\em and} its global depth estimate $\bd^{G\ua}$, we obtain the locally-refined depth value $d_{j}^{R\ua}$ at pixel $j$ as $d_{j}^{R\ua} = \bt^{\ua} \cdot \bphi_j^\ua(\bx)$. Finally, we take this depth estimate at the intermediate resolution ($P_I$ pixels), resize it to the full resolution ($R \times C$) using bilinear interpolation and apply once more the colorization scheme, in order to render the final output more spatially coherent.

It is important to note that besides the use of {\em local} information (rather than the context from the full image), another fundamental difference between our global depth estimation and the refinement lies in the fact that the latter directly regresses on depth, while the former predicts depth reconstruction weights (i.e., the vector $\bw$). This is consistent with the distinct objectives of the two steps: the global estimate takes advantage of the basis constraint to yield a robust but coarse estimate of the depth map; the local refinement can leverage the global depth estimate as a strong feature and thus can model the depth at individual pixels in an unconstrained fashion.

\subsection{Implementation details}
\label{implementationdetails}

In this section we provide additional implementation details concerning our approach. To learn the global depth model, we downsample the training depth maps from size $427 \times 561$ to size $32 \times 43$ for NYUv2 and from $256 \times 1242$ to $32 \times 156$ for KITTI (in order to maintain aspect ratio of ground truth) via bilinear interpolation. The sizes were chosen to reduce the dimensionality sufficiently so as to allow training of basis to happen without overfitting while at the same time producing a coarse depth map that contains meaningful information. Furthermore, we subtract the per pixel mean from each of the depth maps so as to force our model to predict the deviations from the mean depth map. At inference time we add the mean depth to our depth estimate to get the final prediction. As intermediate resolution $P_I$, we use $128 \times 172$ for NYUv2 and $64 \times 311$ for KITTI. 

In the work by Krizhevsky et al.~\cite{Krizhevsky} the deep network was applied to multiple crops of the image and the predictions on the individual crops were then averaged. Inspired by this approach, we defined five distinct image crops (Center (C), Upper Left (UL), Upper Right (UR), Down Left (DL), Down Right (DR)) of size $227 \times 227$ and we learned a distinct global model for each of the crops. However, note that all 5 models are trained to predict the complete depth map (thereby estimating also depth at pixels not in the crop). At inference, we generate the final depth at each pixel as a weighted average of the predictions from the 5 crops. We use a spatially-varying weighting function of the 5 estimates at the coarse size. The weight of crop $i$ at pixel location $p$ is computed as $\beta_i(p) = \exp(-||p-p_i||/\gamma^2) / \sum_{j=1}^5 \exp(-||p-p_j||/\gamma^2)$ where $p_i$ is the center pixel of crop $i$. Thus, at each pixel we give more importance to the predictions of crops that are closer to the pixel.

For each crop, we form the vector centers $\bc_j$ used in the radial basis functions $\phi_j(\bx)$ by taking image examples from the two nearest crops. We use (UL,UR) as centers for C, (C,UR) as centers for UL, (C,UL) as centers for UR, (C,DR) as centers for DL and (C,DL) as centers for DR. {We double the number of centers by including also the mirrored version of each crop in the kernel vector. For NYUv2, as the number of training examples is small ($795$) we use all training images as centers. Thus the RBF kernel vector of each crop contains a total of $795 \times 2 \times 2 = 3180$ centers (mirrored and un-mirrored version of each of the 2 closest crops for all 795 images). For KITTI, since the training set is in this case much larger ($19,852$ images), we use only a subset of it to create the RBF vector: specifically, for each crop we form the centers with the $654$ examples that were used to train the framework of Saxena et al.~\cite{saxena2009make3d}, once again by choosing the mirrored and unmirrored versions of the 2 closest crops for all these images (this yields a total of $654 \times 2 \times 2 = 2616$ centers for each crop).} The $\sigma_{j}$ in the kernel is set to be half of the maximum pairwise distance between centers.

For refinement, instead of learning a single shared model for all pixels of the image, we trained a separate pixel-based model for each block of 16 rows of the image (for a total of $8$ distinct models). This is motivated by the observation that pixels within a row (or in neighboring rows) of the image tend to have similar depth statistics but pixels coming from distant rows often exhibit large depth variations, as already noted in Saxena et al.~\cite{saxena2005learningdepth}. This is merely a consequence of ceilings being typically at the top of the image, walls in the middle and floors at the bottom of the picture. Each model is trained with a $512$ dimensional RBF kernel-vector augmented with the global depth estimate $d_{ij}^{G\ua}$ and the constant feature $1$.  The 512 RBF centers for each block of rows are the k-means cluster centroids obtained by clustering randomly sampled pixels from that block of rows in the training set. Since using the global depth estimates on the training set would overfit the data and generate biased estimate of the feature $d_{ij}^{G\ua}$, we performed a 10-fold cross validation on the training set and used the global depth estimates predicted on each validation fold to generate the features for the subsequent training of the refinement. For each fold, we apply the procedure of training 5 different crops and merging outputs. For colorization, we set the penalty value to 0.001 for both GCL \& RCL.

\section{Experiments}

In this work we apply our proposed approach to the NYUv2~\cite{Silberman:ECCV12} and KITTI~\cite{Geiger2013IJRR} datasets and show that it produces state-of-the-art results on depth estimation for both. These two datasets are dramatically different and serve well the objective of showing that our approach works for both indoor and outdoor settings.

Depth estimation can be quantitatively assessed according to different criteria.
In this work, we report results on multiple metrics that are widely used: RMSE~\cite{EigenPF14}, Absolute Relative error~\cite{EigenPF14}, Scale Invariant error~\cite{EigenPF14}, Threshold error~\cite{Ladicky_2014_CVPR}, Log10 error~\cite{saxena2009make3d}. For evaluation the output of both GCL and RCL is upsampled to full resolution before evaluation. This allows us to compare the global and the refined estimates on the same ground. Both RMSE and Log10 are measured in meters.

The rest of this section is organized as follows: in \S\ref{expNYUv2} we present results of our models on NYUv2 and compare them to the state-of-the-art; in \S\ref{expKITTI} we discuss our experiments on KITTI; finally, in \S4.3 we describe experimental results obtained by varying our model design choices, thus providing further empirical justification for our approach and the settings used in \S\ref{expNYUv2} and \S\ref{expKITTI}.

\subsection{NYUv2}
\label{expNYUv2}
The NYUv2 dataset~\cite{Silberman:ECCV12} consists of RGBD examples from $27$ different indoor scene categories taken from a total of 464 different scenes. 
We evaluate our methods using the standard train/test split provided by the authors of NYUv2 (795 training examples, 654 test examples)~\cite{Silberman:ECCV12}. For the global method we use a depth basis $B$ consisting of $m=48$ atoms. This provided a nice compromise between being able to approximate global depth and predictability of weights of the basis. 


\begin{table*}[!htbp]
{\footnotesize
    \begin{center}
    \begin{tabular}{|c|c|c|c|c|c|c|c|c|c|c|}
	\hline
    	& Metric & GCL & RCL &~\cite{BaigJPBDS14} &~\cite{Ladicky_2014_CVPR} &~\cite{Karsch:TPAMI:14} &~\cite{saxena2009make3d} &~\cite{Liu_2014_CVPR} &~\cite{CVPR15bFayao} & Mean Prediction\\

        \hline\hline
        Higher Better & Th $\delta < {(1.25)}$     							&  0.6083 & 0.6096 & 0.5179  & 0.5422 	& NR 		& 0.447 	& NR 		&  {\bf 0.614} &0.4284 		\\

        \hline
        \multirow{4}{*}{Lower Better}  & Rel     															& 0.2523 & 0.2415 &  0.2544  & NR		 	& 0.374 	& 0.349 	& 0.335 	& {\bf 0.230} &0.4017	 	\\
        & Log10    														& 0.0973 & 0.0960 & 0.1179  & NR 			& 0.134 	& NR 		& 0.127 & {\bf 0.095} 	& 0.1444		\\
        & Sc-Inv   														& 0.2382 & {\bf 0.2363} & 0.2719  & NR 			& NR			& 0.325 	& NR 		& NR & 0.3052 		\\
        & RMSE  															& 0.8156 & {\bf 0.8025} & 0.9917  & NR 			& 1.12 	& 1.214	& 1.060 	& 0.824 & 1.2049	  	\\
        \hline
    \end{tabular}
    \end{center}
    }
\caption{Quantitative Evaluation on NYUv2. Our models (GCL and RCL) outperform prior work by a large margin according to the RMSE metric. Our refinement (RCL) provides a small but consistent improvement over our global estimate (GCL). NR stands for not reported. Results on Make3D were taken from the evaluation of Eigen et al.~\cite{EigenPF14}. Results for Karsch et al.~\cite{Karsch:TPAMI:14} were taken from evaluation with correct train/test split done by Liu et al.~\cite{Liu_2014_CVPR}. The method in Ladicky et al.~\cite{Ladicky_2014_CVPR} was trained with a different train/test split (725 Training, 724 Testing).} \vspace{-0.5cm}
 \label{tb:ResultsNYUV2}
\end{table*}


We compare our approach on this benchmark with published state-of-the-art methods. 
 The results are summarized in Table~\ref{tb:ResultsNYUV2}. The last column reports the performance obtained by simply predicting the constant average depth map (computed from the training set) for any input, as this is an interesting baseline revealing the difficulty of the dataset. As can be seen, both our models outperform all prior methods by a large margin on the RMSE (the metric we optimize for) and are highly competitive with other approaches according to the other performance measures. 

Note that we did not include in Table \ref{tb:ResultsNYUV2} the results of the  method recently proposed by Eigen et al.~\cite{EigenPF14} and Wang et al.~\cite{Wang_2015_CVPR} as these approaches were not trained on the standard training split of NYUv2. Both of these approaches use a training set that is 150 times larger than the one we employ in this work (only 795 images). The results for training with the expanded training set can be seen in Table \ref{tb:largedataset}~\footnote{The results for Wang et al.~\cite{Wang_2015_CVPR} are different from what they report in their paper as they employ a non-standard evaluation method. We used depth estimates provided by the authors and ran our evaluation method to produce the results reported here and in the supplementary material.}. 

\begin{table}[!ht]
{
	\footnotesize
    \begin{center}
    \begin{tabular}{|c|c|c|c|c|c|c|}
	\hline
    	& &  ~\cite{Wang_2015_CVPR}  &~\cite{EigenPF14}& RCL-E  \\
	\hline    	      
        \hline
        Higher Better& Th $\delta < 1.25$     			 &  0.6170 &	0.611	& {\bf 0.6294} \\
        \hline\hline
        \multirow{3}{*}{Lower Better} & Rel     									& {\bf 0.2289 }	& {\bf 0.215} & 0.2250 \\
        & Sc-Inv   								& {\bf 0.228} 	&{\bf 0.219} & 0.2290 \\
        & RMSE  								    & 0.8371 & 0.907 & {\bf 0.7846}  \\        
        \hline
    \end{tabular}
    \end{center}
    }
\caption{Side-by-side comparison between the deep learning based methods~\cite{EigenPF14,Wang_2015_CVPR} and our approach on the test set of NYUv2. RCL-E refers to our model learned on the the expanded training set (the same used by the other approaches~\cite{Wang_2015_CVPR,EigenPF14}).}
 \label{tb:largedataset}
\end{table}  

\subsection{KITTI}
\label{expKITTI}
The KITTI dataset is an outdoor scene dataset consisting of videos taken from a driving vehicle with depth provided by a LiDaR sensor. On this dataset we used the train/test split proposed by Eigen et al.~\cite{EigenPF14} consisting of $19852$ training examples and $697$ test examples. The training and test sets include examples from the ``city'', ``residential'' and ``road'' sequences. For evaluation on this dataset, we use the same experimental setup adopted by Eigen et al.~\cite{EigenPF14}.


We train our global depth model using a basis $B$ consisting of $m=96$ atoms. 
Once again, we compare our estimates against the ground truth by resizing our estimates to full resolution.

Table~\ref{tb:ResultsKITTI} shows the results of our global model versus Eigen et al.~\cite{EigenPF14} (because KITTI is a recent dataset we could not find any other prior work using this training/test split to include in the comparison).
The Table shows that given the same training data, our approach achieves  higher accuracy according to the RMSE and the Threshold metric, while it is  close to the approach of Eigen et al.~\cite{EigenPF14} on the Relative and Scale-Invariant metrics.\vspace{-.1cm}

\begin{table}[!ht]
{\footnotesize
    \begin{center}
    \begin{tabular}{|c|c||c|c||c|c||c|c|}

	\hline
	  & Metric & \multicolumn{2}{|c||}{Coarse} & \multicolumn{2}{|c||}{Refinement} & \\
	  \hline
    	& & GCL  &~\cite{EigenPF14}  & {RCL} &~\cite{EigenPF14} & \begin{tabular}{@{}c@{}}Mean \\ Predict.\end{tabular} \\
	\hline
        \hline
        \begin{tabular}{@{}c@{}}Higher \\ Better\end{tabular} &  \begin{tabular}{@{}c@{}}Th \\ $\delta<1.25$\end{tabular}     			& {\bf 0.691} &  0.679 & {{\bf 0.699}} &	0.692 & 0.556	\\
        \hline\hline
        \multirow{3}{*}{\begin{tabular}{@{}c@{}}Lower \\ Better\end{tabular}} & Rel     									& 0.218  & {\bf 0.194 }	& {0.206}  & {\bf 0.190} & 0.412\\
        & Sc-Inv									& 0.262 &  {\bf 0.248} & {0.260} & 	{\bf 0.246} & 0.359\\
        & RMSE  								    & {\bf 6.608}  & 7.216 & {{\bf 6.437}}  & 7.156 & 9.635\\
        \hline
    \end{tabular}
    \end{center}\vspace{-.2cm}
    }
    \caption{Quantitative Evaluation on the KITTI~\cite{Geiger2013IJRR} dataset. Evaluation was conducted on the test set proposed by Eigen et al.~\cite{EigenPF14} .}\vspace{-0.4cm}
 \label{tb:ResultsKITTI}
\end{table}

\subsection{Revisiting Model Design Choices} \label{modelassessment}

\subsubsection{Global Estimation}

In this section we study the impact of various design choices made in our global approach. Table~\ref{tb:coarseExp} summarizes this comparative study of different variants of our global model (GCL) on both NYUv2 as well as KITTI. 

In this work we assumed that in order to capture the structure in the output space (depth spatial smoothness, rejection of unlikely depth maps), it is beneficial to learn to predict reconstructive depth basis weights rather than regressing on depth directly. The second column of Table~\ref{tb:coarseExp} (Direct Regr) shows the performance obtained by learning a mapping that uses our kernel-based image features $\bphi(\bx)$ to directly regress on the depth $\bd$. As can be seen eliminating the basis model and regressing depth directly causes an increase in RMSE error on both datasets thereby validating the need for a depth basis. 

Another assumption in our approach is that coupling the learning of the basis and the regression provides a beneficial effect as it allows the method to optimize the depth representation for accurate prediction. Our hypothesis is confirmed by the results shown in the third column of Table~\ref{tb:coarseExp} (Uncoupled), which reports the performance obtained by learning a dictionary via sparse coding and then regressing on the weights of the dictionary. There is a clear degradation in accuracy on both datasets when the modeling of depth and the regression optimization are uncoupled.

Finally, we assess which deep features are effective at predicting depth. We consider two types of features, both extracted from the same deep network architecture~\cite{Krizhevsky} but trained on two different datasets: GCL uses "pool5" features trained on PLACES~\cite{zhou2014places}, while GCL-I (last column of Table~\ref{tb:coarseExp}) uses "pool5" optimized for object class recognition on Imagenet~\cite{imagenet_cvpr09}. Our results show that features learned for scene classification perform much better on depth estimation of scenes compared to features trained for object classification.

 \vspace{-.1cm}\subsubsection{Local Depth Refinement}

Here we present experiments that shed light on the role of different components of our local depth refinement (RCL).

First, we assess the advantage of training separate models for different row-blocks of the image. As discussed, for RCL we subdivided the image into 8 non-overlapping blocks of 16 rows and trained a distinct model for each block. We now take a look at the impact of using a single model as opposed to the multi-model setting. In order to construct an equally powerful single model, we construct a $8\times512$-sized RBF descriptor to train the single-model regressor. However, we found that this yields consistently inferior results compared to the multi-model, e.g., the RMSE on NYUv2 is $0.8213$ versus the $0.8025$ of RCL.

%


 In order to show the importance of estimating the global depth before the local refinement, we tried training a variant of RCL that does not include the global estimate $d_{ij}^{G\ua}$ in the feature vector of Eq.~\ref{eq:localfeature}. Effectively this model uses only the local hypercolumn vector to directly regress the depth of each pixel. This results in dramatically worse accuracy: the RMSE error on NYUv2 is 1.1211 instead to $0.8025$! This furthers validates our belief that a good method for local depth estimation requires a really strong global model used as a guidance for further refinement.

\subsection{Analysis of Computational Cost}

We now show that our approach is both scalable and extremely fast to train. We compare the computational cost of our approach to that of other competing methods~\cite{EigenPF14,Wang_2015_CVPR,CVPR15bFayao}. The deep learning approach described in Liu et al.~\cite{CVPR15bFayao} requires $33$ hours for training with a GPU using  the standard training set (795 examples). In contrast, our global framework (GCL) requires approximately $15$ minutes for feature extraction of the standard train/test split NYUv2 dataset and $10$ minutes for learning all $5$ models on a Xeon E5 CPU. The systems described in Eigen et al. \& Wang et al.~\cite{EigenPF14,Wang_2015_CVPR} use $136,847$ and $200,000$ training examples. The training of the coarse model in Eigen et al.~\cite{EigenPF14} takes $38$ hours, while the model in Wang et al.~\cite{Wang_2015_CVPR} takes $4$ days to train using GPUs. The training of the GCL model on the expanded training set ($136,847$ examples) takes only $8$ hours. GCL inference takes place in under a second.

For refinement, the model by Eigen et al.~\cite{EigenPF14} takes $26$ hours for training. In comparison our refinement method (RCL) requires $\frac{1}{2}$ hour to be trained (including the time needed to run k-mean for the RBF centroid computations). We train the independent models in parallel using a cluster, which makes the total training still $\frac{1}{2}$ hour. RCL inference takes $8$ seconds per image.


\begin{table}[t]
{\footnotesize
\begin{center}
\begin{tabular}{|c|c|c|c|c|c|}

\hline
 & GCL & \mbox{Direct Regr} & Uncoupled & GCL-I\\
\hline
NYUv2  									& {\bf 0.8156} & 0.8384 & 0.8843 & 0.8908\\
\hline
KITTI  									& {\bf 6.6078} & 6.7414 & 6.7138 & 6.9923\\

\hline
\end{tabular}
\end{center}\vspace{-.1cm}
}
\caption{RMSE error for different variants of our global estimation method on NYUv2 and KITTI. GCL is our framework from Section 3.1. ``Direct Regr'' uses the image features to directly regress on depth (no basis learning). ``Uncoupled'' learns the basis via sparse coding and then trains a regression model on the learned weights using our features $\bphi(\bx)$. GCL-I corresponds to the use of Imagenet~\cite{Krizhevsky} (rather than PLACES) image features.} \vspace{-.5cm}
\label{tb:coarseExp}
\end{table}

\section{Conclusion}

We presented a novel approach to depth estimation from single image that naturally integrates global and local information. Global cues in the form of deep convolutional features are used to predict the global depth map. In a subsequent stage the estimated global depth map is used to guide local refinement at a higher resolution. Global estimation is formulated as the joint learning of a depth basis and a regression mapping from the image space to the basis weights. The local refinement regresses directly on pixel depth using the global estimate as feature. Our approach yields an improvement over the state-of-the-art on the standard train/test split of the NYUv2 and KITTI datasets. Furthermore it is significantly faster and more scalable than prior systems. Future work will involve integrating feature learning in the framework of coupled regression and modeling of depth.

\section{Acknowledgements}

We thank Loris Bazzani for helpful discussions. This research was funded in part by NSF award CNS-1205521. We gratefully acknowledge NVIDIA for the donation of GPUs used for portions of this work.


{\small
\bibliographystyle{ieee}
\bibliography{new_bib_haris}
}

\end{document}